# Towards Probabilistic Inference of Human Motor Intentions by Assistive Mobile Robots Controlled via a Brain-Computer Interface

Xiaoshan Zhou, SM. ASCE[1] Carol M. Menassa, F. ASCE,[2] and
Vineet R. Kamat, F. ASCE[3]

[1]Ph.D. Candidate, Dept. of Civil and Environmental Engineering, University of Michigan，Ann Arbor, MI, 48109-2125. Email: xszhou@umich.edu
[2]Professor, Dept. of Civil and Environmental Engineering, University of Michigan, Ann Arbor, MI, 48109-2125. Email: menassa@umich.edu
[3]Professor, Dept. of Civil and Environmental Engineering, University of Michigan, Ann Arbor, MI, 48109-2125. Email: vkamat@umich.edu

## ABSTRACT

Assistive mobile robots are a transformative technology that helps persons with disabilities regain the ability to move freely, which in turn empowers them to access public systems and actively engage in social and community life, promoting inclusion and independence. Although autonomous wheelchairs significantly reduce user effort, they still require human input to allow users to maintain control and adapt to changing environments. This input is necessary because current algorithms cannot fully handle all real-world complexities or perfectly match the user's evolving needs. Among different methods for controlling devices (e.g., joysticks, touchscreens, voice commands, or gestures), Brain-Computer Interface (BCI) stands out as a highly effective and user-friendly option. They do not rely on physical movement, making them particularly useful for those with limited hand or arm mobility; they operate silently, maintaining user privacy and minimizing social intrusiveness, and are not affected by external disturbances like loud noise or poor lighting conditions. Current BCI systems can understand whether users want to accelerate or decelerate based on brain signals (measured by electroencephalogram, EEG), but they implement these changes in discrete speed steps rather than allowing for smooth, continuous velocity adjustments. This limitation prevents the systems from mimicking the natural, fluid speed changes seen in human self-paced motion. The authors aim to address this limitation by redesigning the perception-action cycle in a BCI-controlled robotic system: improving how the robotic agent interprets the user's motion intentions (world state) and implementing these intentions in a way that better reflects natural physical properties of motion, such as inertia and damping. The scope of this paper focuses on the perception aspect. We asked and answered a normative question "what computation should the robotic agent carry out to optimally perceive incomplete or noisy sensory observations?" Empirical EEG data were collected, and probabilistic representation that served as world state distributions were learned and evaluated in a Generative Adversarial Network framework. The probabilistic approach not only infers the direction of velocity changes but also quantifies the confidence in these inferences, which is then used to adjust the magnitude of the velocity increment (proportional to its certainty). The ROS framework was established that connected with a Gazebo simulation environment containing a digital twin of an indoor space and a virtual model of a powered robotic wheelchair. Signal processing and statistical analyses were implemented to identity the most discriminative features in the spatial-spectral-temporal dimensions, which are then used to construct the world model for the robotic agent to interpret user motion intentions as a Bayesian observer. This robotic perception module for human intent recognition based on partially informative sensation inputs is critical in various human-robot interaction scenarios, and will be integrated to future testing and validation pipelines as part of our ongoing efforts in a user study focusing on shared control in navigation.

**INTRODUCTION**

Mobile robots, originally developed as Automated Guided Vehicles (AGVs) to improve productivity and operational efficiency in industrial systems, have now also evolved into critical assistive technologies that restore independent mobility for persons with disabilities (PWD) (Al-Qaysi et al., 2018). Specifically, robotic wheelchairs offer enhanced autonomy for individuals with mobility impairments, enabling them to move freely within their homes and social environments, integrate into community life, and regain sense of self-dignity (Lakas et al., 2021). Nevertheless, the sensors and algorithms successfully applied in AGVs—designed for structured environments like Industry 4.0 assembly lines and warehouses—fall short when it comes to assistive mobile robots (AMRs). AMRs must navigate unstructured, real-world scenarios and therefore cannot fully eliminate the need for human input to ensure safety and accommodate personalized situational needs. This research focuses on leveraging Brain-Computer Interface (BCI) to develop highly user-friendly AMRs that cater to individuals with a wide range of mobility limitations (from partial motor function to complete immobility). By directly translating brain signals into robotic commands, BCIs enable discreet operation, preserve social privacy, and are less susceptible to environmental disturbances.

Existing BCI-controlled robotic wheelchair systems can be broadly classified into two categories based on signal types: spontaneous signals and evoked signals. Spontaneous signals are self-initiated brain activities, particularly during motor imagery, where users mentally simulate actions (e.g., moving left or right hand) (Huang et al., 2019). These signals are established and mapped to predefined options for motion control. Evoked signals, on the other hand, are elicited by external stimuli. The best-known examples are P300 (He et al., 2017) and Steady-State Visual Evoked Potentials (Long et al., 2012), which are triggered by flicking stimuli at different frequencies. Given the ability to differentiate between user-desired options, control mechanisms can be categorized into high-level control and low-level control.

In high-level control, users select destinations (Lakas et al., 2021) or trajectories generated by robotic systems (Hamad et al., 2017), and the robot autonomously navigates to the selected goal. This approach significantly reduces user efforts but also limits user flexibility and agency in motion control, potentially undermining the user's sense of safety and comfort—particularly for PWD (Udupa et al., 2021). Low-level control, in contrast, interprets user intentions in real time and converts them into commands for robotic actuators. However, in existing mechanisms, whether high-level or low-level, the control available to users are restricted to discrete options. For example, users can choose between commands such as "forward", "left-turn", or "right-turn" (Chen et al., 2022) or select between velocity levels like "high", "medium", or "low". These commands do not allow users to control velocity in a fully self-paced manner.

The limitation is rooted in the errors in interpreting noisy brain signals (typically captured via EEG). These signals are used as observable data to infer human intentions that are inherently internal and inaccessible. Due to noises arising from "BCI illiteracy" phenomenon where users find it difficult to generate consistent signal patterns and measurement errors (influenced by factors such as scalp conductivity), these observations are only partially informative, and the robotic agent must infer the user' intentions based on these noisy sensory inputs. Previous works have typically handled this uncertainty using a softmax operation (maximum likelihood) at the classifier' output. Yet, an output like [0.55,0.45] for binary discrimination obviously contains more uncertainty compared to an output like [0.8,0.2]. In fact, whenever we, as humans, perceive, we are reasoning with probabilities. Our inference is probabilistic, always accompanied by a degree of uncertainty, even if we do not realize it.

As we develop robotic perception capabilities, we draw critical inspirations from Piaget's cognitive development theory, which outlines how children progress from forming concrete schemas to integrating uncertainty and probabilistic reasoning when they get older. Similarly, we aim to embed such high-level uncertainty awareness into robotic perception systems. Building upon this, our goal is to construct and test the optimal method of probabilistic inference—a Bayesian observer—for the robotic agent to interpret the world state (i.e., the human users' motion intention) based on incomplete observations (noisy EEG signals). For the robotic agent, processing electrophysiological data from humans is a new sensation, and due to the sophisticated nature of human cognitive activities, the generative process and the generative model embedded in the dual perceptual processes need clear definitions.

Therefore, the primary goal of the paper is to provide an intuitive understanding of the perceptual inference and how the world model is shaped in the robotic agent in interactions with human users. We introduce the probabilities involved in perceptual inference—the likelihood, the prior, and the posterior—and how they are related through Bayes' formula. We then collected empirical EEG signals and provided more rigorous mathematical treatments in modelling the world state distributions. In addition, given the high dimensionality in EEG observations, identifying the feature dimension that shows the greatest distinguishability between different world states that would be extracted to construct the "world model" is another objective of this study.

**METHODOLOGY**

**Bayesian Inference.** Perception is a form of probabilistic inference: from incomplete and imperfect sensory observations, the brain strives to figure out the state of the world. Among all possible strategies that can be used for a perceptual inference task, the best possible one is called Bayesian inference. It consists of computing the probabilities of each possible interpretation of the observations at its disposal, and then acting in a manner that has the greatest expected benefit (Wei Ji Ma, 2023). The transitions from sensation to perception requires conditional probabilities. Whether we are aware of it or not, we frequently make judgments based on conditional probabilities in daily life. For example, we estimate $p$(it is going to rain | visual information about the sky) and $p$(that's the president's voice | acoustic information from the radio).

**Generative Process and Generative Model.** Humans are endowed with a collection of exquisite sensory organs through which we detect the environment. Yet, we do not primarily care about the sensory input (e.g., the pattern of light wavelengths or acoustic energy) per se (generative process), but rather about the information the input provides about the state of the world (generative model) (see Figure 1). For the robotic agent embedded in the BCI system, its sensation is the patterns of human brain signals measured via EEG, varying in amplitude and time, and the interpretative transition to perception is to infer the most probable state of the world (in our case, the user's velocity-based motion intention).

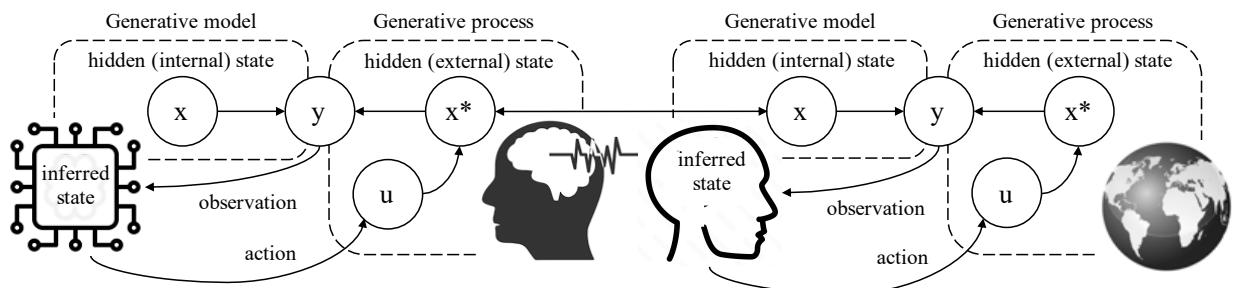

**Figure 1. Generative process and generative model for the BCI (left) and human (right).**

**Conditional Probabilities.** Conditional probabilities are written $p(B\,|\,A)$, and read "the probability of B given A". We refer to A as sensory data, and B as a hypothesized world state. In the proposed BCI system, we are concerned with evaluating the conditional probabilities of just two hypothesized world states ($B_1$: the user wants to speed up, and $B_2$: the user wants to slow down). It is important to note that $p(B\,|\,A)$ is rarely equal to $p(A\,|\,B)$, which is called the prosecutor's fallacy. In fact, Bayes' rules states that

$$p(B_i\,|\,A) = \frac{p(A\,|\,B_i)p(B_i)}{\sum_{k=1}^{N} p(A\,|\,B_k)p(B_k)} \tag{1}$$

We must first consider the sources of the perceptual uncertainty. At the root of the perceptual uncertainty is the fact that different world states can generate the same sensory data, as denoted in $p(A\,|\,B)$. This conditional probability is known as likelihood, which summarizes the degree to which the sensory data favor one world state interpretation over the other. Perceptual inference is based not just on the sensory data (i.e., the likelihood function), but also on expectation.

We represent expectation by prior probability (denoted as *p(B)*, which is the probability of a world state based on all relevant information except the current sensory data. In BCI-controlled wheelchair systems, we know that users would very likely exert specific mental concentration for a short period to serve as a continuous signal indicating their intention, like speeding up. For computational purposes, the brain signals are discretized over time, thus we get a serious of snapshots or intervals. When we assume that the user's intention remains consistent over a short timespan, the prior knowledge will influence how the system infer the user' future intention. For example, if the user is decoded to want to speed up at one time step, the system will assume that they are more likely to maintain this intention in the next. Finally, the brain forms the belief in each possible world state based on sensory data and expectation. The beliefs are called posterior probabilities, as denoted in *p(B | A)* .

**Log Odds and Maximum A Posteriori (MAP) Estimation.** Describing an optimal Bayesian observer is essentially calculating the posterior over A, *p(B | A)*. For the binary discrimination, the posterior distribution is uniquely determined by the ratio of the posterior probabilities of the two alternatives. We can calculate this posterior ratio (or called posterior odds) using Equation (2):

$$\frac{p(B_1 | A)}{p(B_2 | A)} = \frac{\frac{p(A | B_1)p(B_1)}{p(A)}}{\frac{p(A | B_2)p(B_2)}{p(A)}} = \frac{p(A | B_1)p(B_1)}{p(A | B_2)p(B_2)} \qquad (2)$$

In Bayesian calculations, we often see Eq. (2) with the natural logarithm taken of both sides:

$$d(A) = log\frac{p(B_1 | A)}{p(B_2 | A)} = log\frac{p(A | B_1)p(B_1)}{p(A | B_2)p(B_2)} = log\frac{p(A | B_1)}{p(A | B_2)} + log\frac{p(B_1)}{p(B_2)} \qquad (3)$$

The quantity *d(A)* is called the log posterior odds, which is symmetric. If the log posterior odds is positive then *p(B₁ | A)* is larger than *p(B₂ | A)*. When two alternatives have the same probability, the log posterior odds is zero. Therefore, the MAP estimation is determined by the sign of the log posterior odds. The inequality used to determine the MAP estimate, *d(A)>0*, is called the decision rule of the Bayesian MAP observer, and *d(A)* is called the decision variable. The scalar value to which the decision variable is compared in order to make a decision, here 0, is called the decision criterion.

The criterion, 0, is determined by the fact that we adopt a flat prior, meaning we do not make assumptions about the temporal dependencies of time steps but rather assign equal probabilities to all interpretations without any expectation. This is because we think the prior information can introduce biases that may incorrectly favor one interpretation over the other. Consider three consecutive time steps. Initially, when the system detects a signal indicating a desire to speed up, we cannot determine whether this signal represents a deliberate action or an involuntary fluctuation, where the user remains idle but the signal coincidentally falls within the "active" state range. Further, even if the system infers with high certainty that the user intends to speed up during the first two time steps, we still cannot assume that the same intention persists at the third step. The user's intention may change unpredictably, but the system's reliance on temporal dependencies can lead to cumulative bias, where it reinforces an initial assumption, such as speeding up, despite the possibility that the user' intention may change. Therefore, although temporal dependencies aim to smooth velocity transitions and reduce abrupt changes, they can introduce additional safety concerns of "velocity collapse", where the system loses real-time control over velocity due to over-committing to irretrievable cumulative effects.

We use a flat prior and make inferences only based on the incoming sensory evidence at the current time interval. We treat the physical dynamics of velocity changes during the action stage, where a particle filter is designed to create the "viscosity" in velocity change by reconfiguring a large number of particles in the mathematical space where velocity-related attributed are represented. This forms a perception-action closed-loop. These two components are interconnected in the system design, but the scope of this article focuses mainly on perception.

**Inference Confidence.** In a binary decision, the log posterior odds also has a magnitude or absolute value. A lower absolute value means that the posterior probabilities of the two alternatives are closer to each other. Therefore, a natural measure of confidence in a binary decision is the magnitude of the log posterior odds:

$$confidence = \left| log \frac{p(B_1 \mid A)}{p(B_2 \mid A)} \right| = \left| log \frac{p(A \mid B_1)}{p(A \mid B_2)} \right| \qquad (4)$$

This inference confidence is used to determine the magnitude of velocity increments in the BCI-controlled robotic motion system.

**EEG Data (Measurement) Collection.** Having established the perception module of the robotic agent as an optimal Bayesian observer for recognizing human intent, we will follow the terminology of the sensory inputs available to it, i.e., the measurement $A$ from the generative process. These measurements were collected using non-invasive EEG technology, with the Emotiv Epoc X headset containing14 channels (AF3, F7, F3, FC5, T7, P7, O1, O2, P8, T8, FC6, F4, F8, AF4). The headset uses felt pads soaked in saline solution to ensure conductivity between the scalp and electrodes. The raw EEG signals were transformed into the frequency domain using the Fast Fourier Transform (FFT), because EEG oscillatory activity is more informative than raw potentials for evaluating the temporal dynamics of neuronal network activation and coordination across cerebral regions to support sensory, motor and cognitive processing (Deiber et al., 2012). This resulted in a 70 dimensional feature space that includes the 14 channels, each analyzed across five frequency sub-bands: theta (4–8 Hz), alpha (8–12 Hz), low beta (12–16 Hz), high beta (16–25 Hz), and gamma (25–45 Hz). The EEG power data were recorded at a sampling rate of 8 Hz.

**The Likelihood Function.** The optimal Bayesian observer makes perceptual inferences based on the log posterior odds, using knowledge derived from the generative model. As discussed earlier, when the prior is flat (i.e., all hypotheses are equally likely), the decision variable is predominantly determined by the likelihood odds. The likelihood function represents the probability distribution of the observer's belief about how likely a given measurement is under each hypothesized world state. To construct the state distribution, we developed a structured procedure using PsychoPy for experimental design. In this procedure, users are prompted to mentally simulate speed-up and slow-down actions. The detailed settings, including the stimuli configuration, trial-based procedure, stimulus presentation duration, and loop setup in the PsychoPy interface, are illustrated in Figure 2. PsychoPy, an open-source psychology software, was used to deliver the visual stimuli and automatically synchronize EEG data recording, event triggers (temporal markers), and labels with Emotiv Pro. The two action cues are presented in a random order to minimize habituation and anticipation effects in fixed block designs. These temporal markers are critical for providing the pointers to timestamps of stimulus onset, which were then used to segment EEG signals from the continuous recording. Depending on the length of the sliding window, the power within each window is averaged and used to construct the likelihood probability distribution function (PDF).

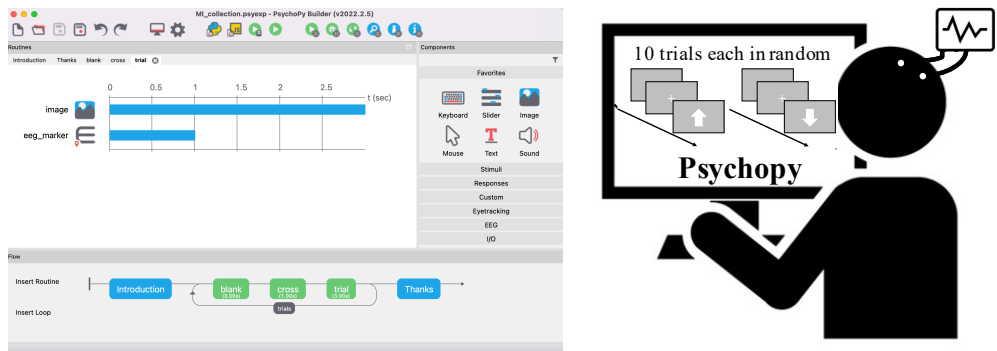

**Figure 2. Experimental design in Psychopy for the likelihood function construction.**

**Robot simulation in Gazebo and Robot Operating System (ROS).** We developed a workflow to subscribe real-time EEG data from the Emotiv Cortex API and run the robotic agent's perception computation in Python using Visual Studio Code. The computed velocity commands are then forwarded to the robotic actuators via ROS bridge (the rospy library). Specifically, the commands (linear velocity) are published as geometry_msgs/Twist messages on the /cmd_vel topic. The robotic platform is a digital replica of a real powered wheelchair, modeled using URDF. The robot's embodiment was modularly mapped for control, and its joint drive strength and damping parameters were fine-tuned for realistic behavior. The environment is a digital twin of an indoor building. Gazebo was used to provide a high-fidelity interface for visualizing the robot's motion in physically accurate scenarios. Additional configuration details are illustrated in Figure 3.

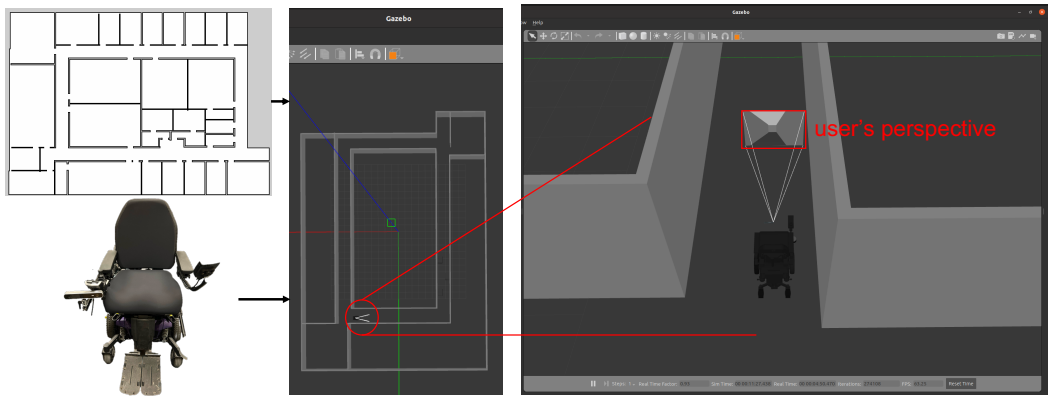

**Figure 3. The Gazebo environment, with a digital replica of a robotic wheelchair testbed and an indoor university building space.**

**RESULTS**

To determine which feature dimensions exhibited the greatest distinguishability between two world states, a Mann-Whitney U test was performed. Features meeting the statistical threshold of $p<0.05$ were further analyzed for effect size (r) and ranked in descending order (see Figure 4). For data segmentation, we applied a sliding window of 1000 ms with a 500 ms overlap, extracting frequency-specific power data and averaging it across each channel. The results indicate that, during the first 1000 ms after stimulus onset, only the right occipital region showed measurable differences between the two states, indicating the primary involvement of visual areas. Beyond 500 ms post-stimulus, we observed broader global activity reconfigurations reflecting motor execution processes, particularly in the alpha and low beta frequency bands. These findings are consistent with existing literature demonstrating that modulations in sensorimotor functions are predominantly associated with the alpha and beta frequency ranges (Deiber et al., 2012).

Further visualization revealed that acceleration was associated with stronger event-related synchronization in regions of the right frontal and temporal areas, compared with the relaxation state recorded from the same user at idle. To differentiate between "idle" and "active" states (actively intending to adjust velocity), we calculated Z-scores exceeding 2 standard deviations from the relaxation state. Results showed that acceleration produced more distinguishable patterns than deceleration. This suggests that exerting deceleration may require additional safety mechanisms (emergency-stop button or multi-modality) in the system design. In practice, we observed that powered wheelchairs often operate at very low speeds with large safety buffers, averaging 6-8 km/h compared to the ISO standard of 15km/h. This motivates us to design a system that empowers users to self-regulate higher speeds when they perceive the environment as safe. It is also important to note that for each individual user, the analytical procedure is conducted independently to personalize feature selection. Features in other significant dimensions displayed similar patterns but we did not foresee increased discriminatory power when using a joint PDF.

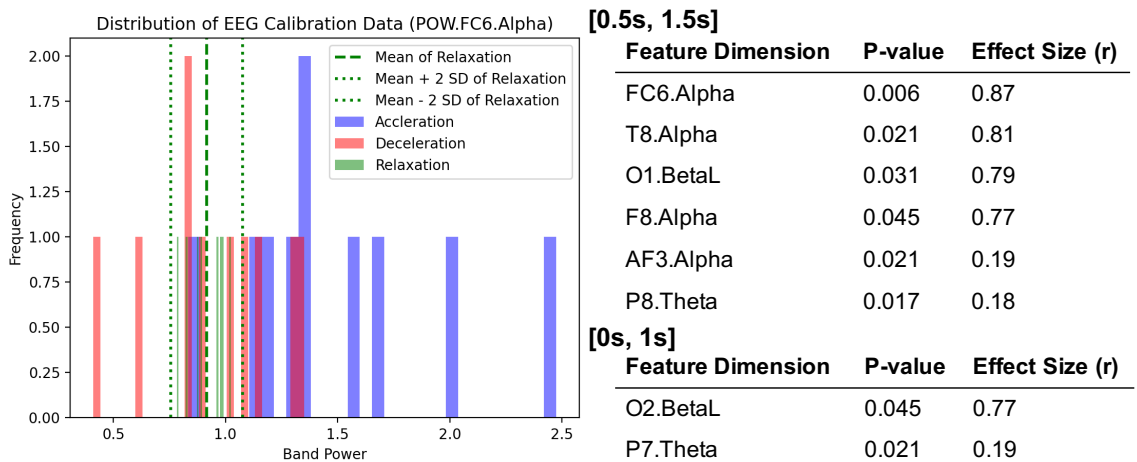

| [0.5s, 1.5s] | | |
|---|---|---|
| Feature Dimension | P-value | Effect Size (r) |
| FC6.Alpha | 0.006 | 0.87 |
| T8.Alpha | 0.021 | 0.81 |
| O1.BetaL | 0.031 | 0.79 |
| F8.Alpha | 0.045 | 0.77 |
| AF3.Alpha | 0.021 | 0.19 |
| P8.Theta | 0.017 | 0.18 |
| [0s, 1s] | | |
| Feature Dimension | P-value | Effect Size (r) |
| O2.BetaL | 0.045 | 0.77 |
| P7.Theta | 0.021 | 0.19 |

**Figure 4. Distribution of POW.FC6. Alpha data and statistical comparison.**

To construct the likelihood PDF, the distributions of the two world states were compared against a set of known distribution shapes (including Gaussian, exponential, logistic, Cauchy, gamma, beta, Weibull, Pareto, chi-squared, Student's t, and Rayleigh distributions) using Kolmogorov-Smirnov tests. Ideally, a Gaussian (normal) distribution would allow for a straightforward mathematical formula to evaluate the PDF for real-time power values. Given the small sample size, the Shapiro-Wilk test was performed to assess the normality of the EEG data. Both world states (acceleration and deceleration) failed to reject the null hypothesis of normality ($p = 0.52$ for acceleration; $p = 0.85$ for deceleration), confirming that the data approximated a normal distribution. In fact, the primary concern is the distinguishability of the two world state distributions, as this characterizes the robotic perception system's ability to discriminate between them. To measure this, the Kullback-Leibler (KL) divergence was calculated, providing an indicator of how distinct the data patterns are. The normalized KL divergence can also serve as a screening criterion for reliably onboarding users into the BCI system. Validating this measure will require a larger user sample in our future endeavors.

To address the challenge of limited trial data—intended to minimize user calibration time—we explored data augmentation techniques to construct more robust PDFs. Regardless of normality, we defined a set of measures to characterize the central tendency and variability of each state distribution, including median, interquartile range (IQR), mean absolute deviation (MAD), skewness, and kurtosis. We implemented a Generative Adversarial Network (GAN) architecture to generate synthetic data points that reflect the characteristics of the original data patterns. Comparisons between the generator's performance using cross-entropy loss and a customized loss function incorporating the five distribution measures (median, IQR, MAD, skewness, and kurtosis) after the same training epochs are shown in Figure 5. This data augmentation approach also has implications for user data privacy, as synthetic data can help anonymize sensitive user information. Further details on privacy considerations will be disclosed in future publications.

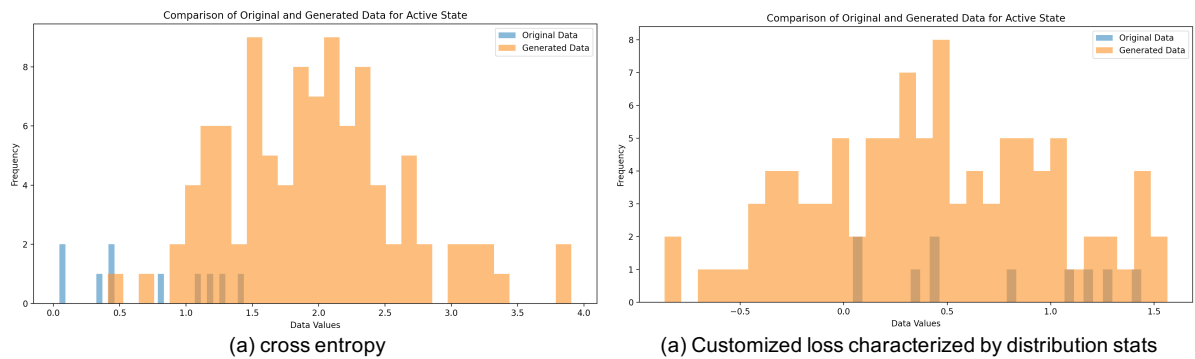

(a) cross entropy     (a) Customized loss characterized by distribution stats

**Figure 5. Distribution of original and synthetic data distributions under two loss functions.**

**CONCLUSION**

This study on a BCI-controlled robotic wheelchair lies at the intersection of a wide range of disciplines, including neuroscience, psychology, computer science, physics, statistics, and engineering. The primary focus of this paper is on enhancing the robotic agent's perception module as an optimal Bayesian observer. This initial work aims to explore the technical feasibility of this idea and the proposed workflow and thus for the technical verification, human data was only collected from individuals in the authors' research group and not from external "users", which is a next logical step in the project. The perceptual output generated by what is described here will be forwarded as velocity commands into the action loop for the robotic actuator. The results from this initial study offer significant promise and the authors aspire to develop more intelligent assistive mobility solutions for vulnerable populations.

## ACKNOWLEDGEMENTS

The authors would like to acknowledge the financial support for this research received from the US National Science Foundation (NSF) via Grant SCC-IRG 2124857. Any opinions and findings in this paper are those of the authors and do not necessarily represent those of the NSF.